\documentclass[10pt, a4paper, table]{article}

\usepackage[]{lrec-coling2024} 

\usepackage{tabularx}
\usepackage{booktabs}
\usepackage{enumitem}

\usepackage{xcolor}
\usepackage{hyperref}
 \definecolor{darkblue}{rgb}{0, 0, 0.5}
  \hypersetup{colorlinks=true, citecolor=darkblue, linkcolor=darkblue, urlcolor=darkblue}

\usepackage{xstring}

\usepackage{color}

\title{Investigating the Robustness of Modelling Decisions for Few-Shot Cross-Topic Stance Detection: A Preregistered Study}

\name{Myrthe Reuver$^1$, Suzan Verberne$^2$, Antske Fokkens$^1$$^,$$^3$} 

\address{
$^1$Computational Linguistics and Text Mining Lab, Vrije Universiteit Amsterdam\\
$^2$Leiden Institute of Advanced Computer Science, Leiden University\\
$^3$ Dept. of Mathematics and Computer Science, Eindhoven University of Technology \\
\texttt{$^1$firstname.lastname@vu.nl}, \texttt{$^2$s.verberne@liacs.leidenuniv.nl}
}           

\abstract{
For a viewpoint-diverse news recommender, identifying whether two news articles express the same viewpoint is essential. One way to determine "same or different" viewpoint is stance detection. In this paper, we investigate the robustness of operationalization choices for few-shot stance detection, with special attention to modelling stance across different topics. Our experiments test pre-registered hypotheses on stance detection. Specifically, we compare two stance task definitions (Pro/Con versus Same Side Stance), two LLM architectures (bi-encoding versus cross-encoding), and adding Natural Language Inference knowledge, with pre-trained RoBERTa models trained with shots of 100 examples from 7 different stance detection datasets. Some of our hypotheses and claims from earlier work can be confirmed, while others give more inconsistent results. The effect of the Same Side Stance definition on performance differs per dataset and is influenced by other modelling choices. We found no relationship between the number of training topics in the training shots and performance. In general, cross-encoding out-performs bi-encoding, and adding NLI training to our models gives considerable improvement, but these results are not consistent across all datasets.  Our results indicate that it is essential to include multiple datasets and systematic modelling experiments when aiming to find robust modelling choices for the concept `stance'.
 \\ \newline  
 }

\begin{document}

\maketitleabstract

\section{Introduction}
Recently, work has stated the importance of different viewpoints in news exposure for a healthy democracy \cite{helberger2019democratic,  mattis2022nudging}. Designing a viewpoint-diverse news recommender requires measuring how news articles differ in viewpoint. One way to determine difference in viewpoint is to measure different stances (pro versus con) towards a topic in recommended news articles, as \citet{alam2022towards} do on German news texts covering the immigration debate. 

However, a challenge for the stance approach in viewpoint detection is that the news continuously has new discussion topics. Models trained on stance detection in one topic do not necessarily work well on other topics \cite{reuver2021stance, jakobsen2021spurious}. Cross-topic performance of stance models is influenced by different choices when modelling the concept "stance": task definition, dataset, modelling architecture and model pre-training knowledge. We 
compare modelling choices for \textit{few-shot cross-topic and in-topic stance detection}.

\begin{figure*}
    \centering
    \includegraphics[width=\textwidth,scale=0.27]{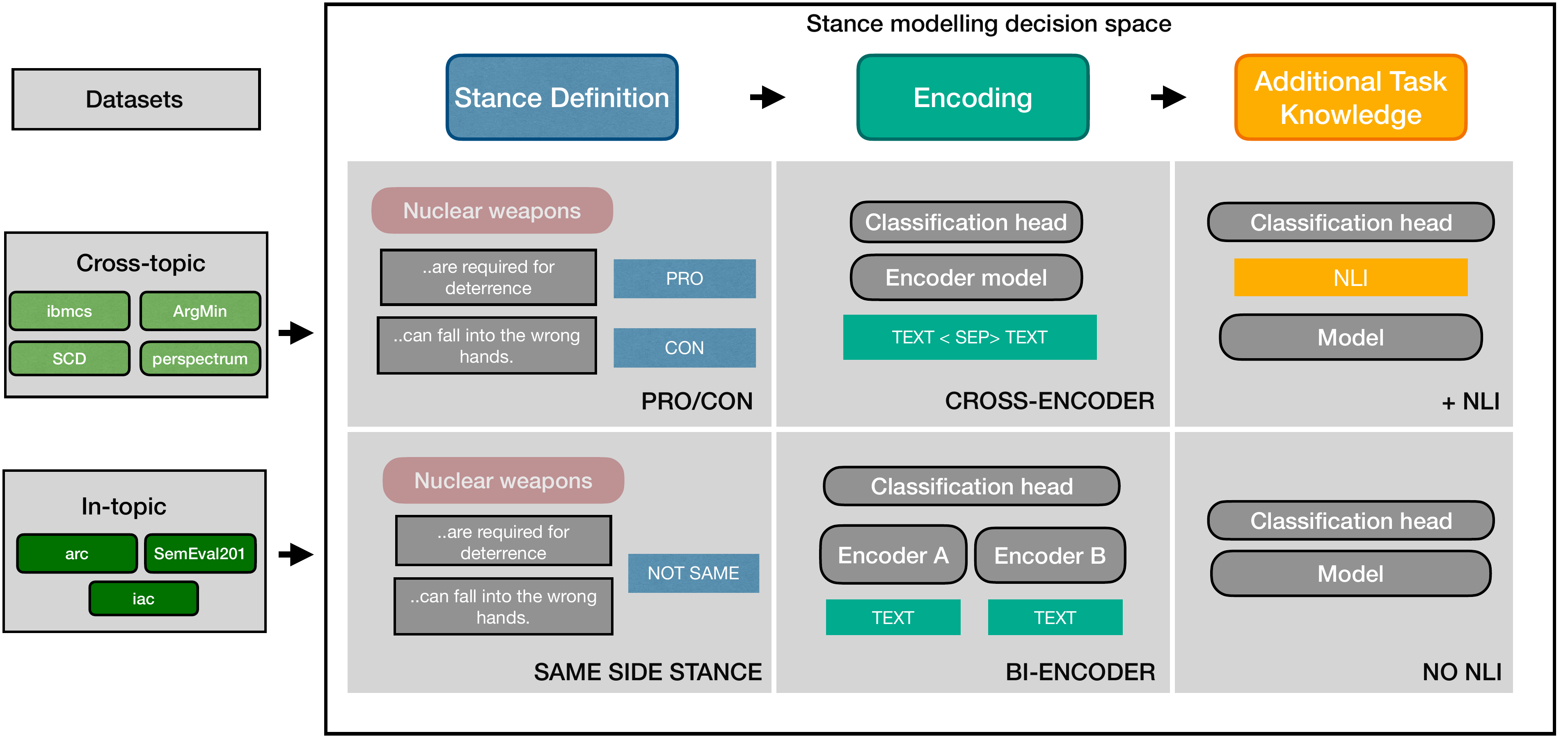}
    \caption{The experimental set-up of this paper, visualizing the different decision points being researched as well as the different scenarios (cross and in-topic) and seven training and evaluation datasets.}
\end{figure*}\label{fig:researchprocess}

One modelling choice is task definition: Stances are usually defined as statements in favour (pro), against (con), or neutral towards a topic or goal \cite{thomas2006get, mohammad2016semeval, kuccuk2020stance, schiller2021stance}. We compare this task definition to Same Side Stance Classification (SSSC) \cite{stein2021same}. SSSC is the task of classifying whether a pair of texts express the same (e.g.\ pro+pro, con+con) or a different (e.g.\ pro+con) stance on a topic. The second modelling choice is the training architecture: 
we compare cross-encoding and bi-encoding. We also investigate modelling with or without knowledge of Natural Language Inference (NLI).

We preregistered potentially fruitful hypotheses and experiments before running experiments. Our preregistration follows the preregistration recommendations by \citet{van2021preregistering}.
In a preregistered report,\footnote{\url{https://osf.io/zrhe7/?view_only=ee2870247e4a47678014b1d7b7b7a943}} we registered our research questions, datasets, and hypotheses before doing our experiments and analyses. We adhered to the question: ``Would a reader of the manuscript wonder whether a given decision about analysis, data source or hypothesis was made after knowing the results?'' \footnote{\url{www.aspredicted.org/messages/why_limits.php}}

We address two research questions:

\begin{itemize}[noitemsep]
\item How do different modelling choices (task definitions and architecture differences) affect few-shot classification performance on different stance datasets?
\item To what extent do these modelling choices affect few-shot cross-topic robustness? 
\end{itemize}

\newpage

This work adds three main contributions to the literature: (1) we investigate the effect of several core modelling choices and their interaction on cross-topic stance modelling, including task definition (Pro/Con versus Same Side Stance); (2) We adapt a benchmark of seven datasets on Pro/Con stance detection into one for also measuring Same Side Stance, and release this adaptation; 
and (3) we are, to our knowledge, the first study to pre-register hypotheses and experiments in computational argumentation. We hope this will inspire other researchers to also pre-register their studies, for transparency and systematicity. For additional transparency, we also release our experimental code.\footnote{\url{https://github.com/myrthereuver/ModelDecisionsStance}}

\section{Background}

We address different modelling choices for stance detection: task definition, architecture, and related task knowledge.  Section~\ref{sec:operat} discusses related work on each of the choices. Section~\ref{sec:Eval} discusses earlier literature on cross-topic stance detection, especially in the few-shot setting, and relates this work to the experimental modelling choices represented in Figure~\ref{fig:researchprocess}.

\subsection{Stance Modelling Decisions
}\label{sec:operat}

\textbf{Stance Definitions}
Previous work has measured different viewpoint in news articles as opinion articles with a different framing of the same issue or topic \cite{mulder2021operationalizing} 
or articles with a different stance score on the same issue \cite{alam2022towards}. Stance is a useful operationalization of viewpoint: it captures different positional opinions on socio-political issues \cite{du2007stance}. There are several different task definitions for stance, which we explain below.

One option is defining stance as pro, con, or neutral towards a topic. Such stance classification of arguments \cite{thomas2006get, mohammad2016semeval, kuccuk2020stance, schiller2021stance} may have limitations for our use-case of viewpoint diversity in different news topics, since it has shown limited cross-topic generalizability \cite{reuver2021stance,jakobsen2021spurious}. 

Therefore, viewpoint diversity may be better operationalized with another task definition for stance: Same Side Stance Classification (SSSC). This task is designed by \citet{stein2021same}, who released a SSSC shared task dataset and argue this task is more robust in the cross-topic scenario.  SSSC is the task of classifying whether a pair of texts express the same (e.g.\ pro+pro, con+con) or a different (e.g.\ pro+con) stance on a topic.  \citet{korner2021classifying} use the shared task dataset with 2 topics to train a cross-topic, cross-encoder RoBERTa model. Their extensive dataset analysis shows that SSSC models still under-perform on unseen debate topics, with task-specific overfitting on certain debate topics. Their deduplication efforts give more realistic cross-topic performance than earlier versions of the shared task data.

More recently, \citet{figueras2023dynamic} present a multilingual stance dataset where pairs of social media comments are annotated as having an agree or disagree relationship rather than each text being separately annotated as pro or con towards a topic. This shows a pair-wise same/different definition of stance may also be relevant to the social media domain.

\textbf{Input Encoding: Bi vs Cross-Encoding}\label{sec:architectures}
Another choice point in modelling chance is choosing a specific architecture and model, specifically: different input encodings.
Stance can be modelled by explicitly modelling similarity of the different stance classes, e.g.\ for \textit{pro vs con} when modelling the Same Side Stance definition. This can be done with bi-encoding approaches such as SBERT \cite{reimers2019sentence}. Bi-encoding approaches encode each input document separately as an embedding representation, and then measure similarity between these representations. Such bi-encoding approaches have successfully been used for pair- or triplet-wise stance detection, often with the topic or claim under discussion incorporated in the input pair or triplet \cite{popat2019stancy,hossain2020covidlies,yang2021tribrid}. This approach is in contrast to 
cross-encoder architecture for text classification in Transformer models, in which two texts are concatenated (i.e.\ [arg1] [SEP] [arg2]), after which a classification head is usually fine-tuned to make a classification decision on the texts.

\textbf{Additional Task Knowledge}
Inter-training or Intermediate Task Training \cite{pruksachatkun2020intermediate} is a form of task transfer: leveraging a model's training in Task A for predicting on Task B \cite{vu2020exploring,Albalak2022FETAAB}. 
A stance-related task is NLI (Natural Language Inference). Similarly to stances, this task also models the relation between texts and a topic or discussion. The task is usually defined as classifying whether texts entail, contradict, or are neutral towards a specific premise \cite{nie-etal-2020-adversarial}. Due to the similarity to the stance detection task and the fact it is a well-resourced task with many large benchmark datasets, NLI has been used before as pre-fine-tuning task for stance classification \cite{hossain2020covidlies,hou2022covmis}.

\subsection{Cross-topic Stance Modelling}\label{sec:Eval}
Training on one topic and then classifying the stance of texts on a different topic is a known challenge for stance detection models. Some previous research has approached this challenge by looking into topic relatedness, either during evaluation or training \cite{xu2018cross,wei2019modeling}. 
Other stance work, especially with pre-trained Transformer models, does not explicitly consider the relatedness of topics when designing or evaluating a cross-topic stance detection approach, i.e.\ as in \citet{reimers2019classification}.  

In the following paragraphs, we discuss previous work on cross-topic stance detection. For each work, we highlight which comparisons of the three modelling choices from Figure \ref{fig:researchprocess} and Section \ref{sec:operat} are investigated. This shows that there is no previous work that comprehensively and directly compares the effect of all these choices for few-shot cross-topic stance detection.  

Earlier reported results highlight the difficulty of cross-topic stance detection with Transformer models. \citet{allaway2021adversarial} consider it a more fair evaluation of stance when one does not consider topic-relatedness, as this truly measures whether the model robustly models stance in different topics. Their adversarial model trained to reduce dependency on topic-specific information achieves a cross-topic performance of $F1 = .49$ and $F1 = .54$ on two unseen topics from the semeval2016 dataset \cite{mohammad2016semeval}. \citet{hardalov-etal-2021-cross} already showed different stance datasets respond differently to cross-domain training on full datasets. Their approach is based on RoBERTa-base with different domain encoders, tested on 16 different stance datasets with pro/con stance labels. Their results across datasets is between $F1 = .32$ and $F1 = .82$. Their experiments thus focus mostly on the encoding as well as added task knowledge, e.g.\ the second and third column in Figure~\ref{fig:researchprocess}, but do not compare stance definitions (the first column).

Recent work indicates that topic-agnostic stance models may not actually use cross-topic knowledge \cite{reuver2021stance, jakobsen2021spurious}. Follow-up work focuses therefore on improving the topical knowledge of stance models, either by increasing the diversity of stance topics in the training data \cite{ajjour2023topic} or building on earlier work that aims to improve Transformer models' use of topic-related contextual knowledge \cite{liu2021enhancing,beck2023robust}. \citet{beck2023robust} use contextual knowledge in a dual-encoder set-up to achieve up to $F1 +.12$ on unseen topics.  These works focus mostly on the data and additional (task) knowledge, e.g.\ the last column in Figure~\ref{fig:researchprocess}, but they do not compare stance definitions (the first column). 

Closest related to our work is \citet{varadarajan-etal-2022-detecting}, who show promising results of cross-topic robustness of the Same Side Stance task definition in the few-shot setting with RoBERTa  - evaluating on 34 different topics from a debate forum dataset. They also test generalization to a Twitter dataset. Their results indicate that only a few examples are enough for a model to generalize Same Side Stance to five unseen discussion topics, when training on as few as four examples from different topics, with best performance when training on 32 different topics of $F1 = .745$. However, rather than our two-class same or different stance task on the same topic, they add a third class with a pair of arguments from unrelated topics. Earlier computational argumentation research has showed that having an `unrelated' or `no argument' class can inflate performance \cite{reuver2021stance}. Additionally, they do not compare the Same Side Stance definition to Pro/Con Stance or the influence of NLI knowledge (the first and last column of the experimental set-up in Figure~\ref{fig:researchprocess}).

\textbf{To summarize:} Earlier cross-topic stance work has either not worked on the few-shot setting, not looked beyond one stance definition (either pro/con or Same Side Stance), or has focused on a small number of topics or datasets. Also, earlier work addressed one choice at the time (definition, data, architecture), 
without direct comparison of how these choices interact. Finally, no previous study into cross-topic stance robustness has pre-registered experiments and hypotheses.
This paper systematically 
investigates how task definitions (pro/con versus same/difference) and input encodings (bi and cross-encoding) as well as related task knowledge (NLI) influence cross-topic robustness in a few-shot setting.

\section{Preregistration and Hypotheses}

Our study set-up (including hypotheses, datasets, and experiments) was preregistered on 26-05-2022, before running the experiments. The preregistration is hosted on the Open Science Foundation (OSF) portal.\footnote{ \url{https://osf.io/zrhe7/?view_only=ee2870247e4a47678014b1d7b7b7a943}} 

We outline the hypotheses tested in this paper below, connected to three choices: definition (Section~\ref{sec:def}), encoding ( Section~\ref{sec:encoding}), and added task knowledge (Section~\ref{sec:task}). These hypotheses and modelling choices are also visualized in Figure~\ref{fig:researchprocess}. We also explain the experiments to test the hypotheses (Section~\ref{sec:exp}), and preregistration changes (Section~\ref{sec:changes}).

\subsection{Task Definition}\label{sec:def}
\textbf{Hypothesis 1.1} 
We expect the SSSC definition to be more cross-topic robust than the pro/con stance task definition. This is motivated by the claim in \citet{stein2021same} that SSSC models can solve stance in a topic-agnostic manner, as a SSSC model leans on similarity within a pair of arguments rather than prior knowledge of the topic.

\textbf{Hypothesis 1.2} 
We expect that the size of the topics in training/test splits does not correlate with the classification performance in cross-topic pro/con stance classification. This was also found in  \citet{reuver2021stance}, who surmised that topic relatedness was much more important.

\subsection{Encoding Choices}\label{sec:encoding}
\textbf{Hypothesis 2.1} 
When comparing bi-encoding to cross-encoding in the Same Side Stance task, we expect bi-encoding to fluctuate less between in-topic to cross-topic performance, and improve cross-topic performance. We expect this because the bi-encoding architecture focusses on difference and similarity between two texts. The expected cross-topic stability is partly based on earlier claims on Same Side Stance \cite{alshomary2021siamese, stein2021same}. 

\textbf{Hypothesis 2.2} 
We expect cross-encoding to perform better in both cross-topic and in-topic the held-out test sets. This would be in line with the shared task results in \citet{stein2021same}.

\subsection{Task Knowledge}\label{sec:task}

\textbf{Hypothesis 3.1} 
Because of the extra knowledge added to the model by the additional training, we expect that adding NLI training to the model will lead to classification performance gains over models without NLI training in pre-fine-tuning. This is based on recent argument mining tasks where pre-fine-tuning on NLI improved performance \cite{van2022will, heinisch2022overview}.

\subsection{Experiments to test our hypotheses}\label{sec:exp}

We test the influence of (1) task definition (pro/con versus same/different stance),  (2) architecture choices (bi vs cross-encoding) and (3) pre-fine-tuning on NLI on stance classification performance in the within-topic and cross-topic scenario. We test our hypotheses on training samples of 100 examples from 7 different datasets in the stance benchmark \cite{schiller2021stance}.

\subsection{Changes to the preregistration}\label{sec:changes}
Some pre-registered hypotheses we leave to future work, such as hypotheses on different pre-trained models and clustering as pre-fine tuning task \cite{shnarch2022cluster}.
We chose RoBERTa \cite{liu2019roberta} as our base model for all model comparisons rather than comparing different language models, to specifically analyze the effect of different \textbf{choices} rather than different pre-trained \textbf{models}. Our focus also shifted more to few-shot rather than full dataset learning. We also left out additional datasets, 
because within the stance benchmark there are already sufficient interesting contrasts: of within-topic or cross-topic datasets, as well as different stance labels. Additionally, rather than analyzing results as averages over modelling choices, we decided to mostly look into patterns between modelling choices and datasets, and differences between datasets.

\section{Experimental set-up}

\subsection{Data}
\textbf{Stance Benchmark} 
\citet{schiller2021stance} unified 10 English-language datasets in one large pro/con stance benchmark. The 10 different stance datasets are from different domains (social media, debate forums, news) and on different topics (e.g.\ abortion, hydro-electric dams, male infant circumcision). This benchmark consists of 99,224 training examples, 17,938 development examples and 43,944 of test examples, and 6,168 topics.

\begin{table*}[t]
\centering
\small{
\begin{tabular}{l|rcccccccc}
\toprule
&  & \textbf{ibmcs} & \textbf{pers} & \textbf{argmin} & \textbf{iac1} & \textbf{scd*} & \textbf{arc} & \textbf{sem2016}\\
 \midrule
labels & &  2 & 2 & 2 & 3 & 2 & 4 & 3 \\
  \midrule 
domain &  &  \multicolumn{6}{c|}{forum} & Twitter\\
 \midrule
Evaluation & \multicolumn{5}{c|}{cross-topic} & \multicolumn{3}{c}{in-topic} \\
\midrule
SSSC-100-train topics  & & 21 & 80 & 5 & 6 & 3\textsuperscript{*} & 64 & 5 \\
ProCon-100-train topics & &  22 & 80 & 5 & 6 & 6\textsuperscript{*} & 69 & 5 \\                        
\midrule
Topics test &  & 30 & 227 & 2  & 2 & 4\textsuperscript{*} & 183 & 5\\
Topics same in train/test & &  0 & 0 & 0 & 0 & 4\textsuperscript{*} & 183 & 5\\
\midrule
SSSC test & &1154 & 2230 &  2571 & 762 & 67\textsuperscript{*} & 414 & 699 \\
Pro/Con test & & 1355 & 2773 & 2726 & 924 & 964\textsuperscript{*} &  3559 & 1249 \\
\bottomrule
\end{tabular}
}
\caption{\small{Summary statistics of the benchmark datasets. 
*The SCD dataset has implicit topics, we match with semi-automatic word list on four topics: Obama, Marijuana, Gay Marriage, and Abortion.
}}
\label{tab:datasets}
\end{table*}

\textbf{Selection of Datasets from the Benchmark} 
 Our research aim is on stances towards societal topics, in order to allow for diverse viewpoints on public discussions. 
Therefore, we remove the 3 of the 10 benchmark datasets that contain stances towards other texts, such as headlines, rumours, or tweets, only including the datasets with stance towards a broader societal discussion topic or proposal. This led to a selection of 7 datasets: \textit{arc} \citeplanguageresource{arc},
\textit{semeval2016} \citeplanguageresource{mohammad2016semeval}
\textit{argmin} \citeplanguageresource{argmin},
\textit{iac} \citeplanguageresource{walker},
\textit{scd} \citeplanguageresource{data3},
\textit{ibmcs} \citeplanguageresource{data4} and
\textit{perspectrum} \citeplanguageresource{data5}. Many of these are also used by previous work \cite{hardalov-etal-2021-cross, arakelyan-etal-2023-topic}. The datasets differ in types of texts and topics as well as number of topics, though some datasets have the same topics (e.g.\ gun control, abortion).\footnote{Topics per dataset can be found in our GitHub repository.} We use a random set of 100 examples from each dataset for training, and kept the test sets as-is. These training samples are representative in number of topics.

\textbf{Same Side re-formulation}
We filter out examples with labels that are not pro or con (e.g.\ `neutral', `commenting', or `unknown' stance), and re-code each of the datasets into pairs of two texts. The pairs are then labelled TRUE if both texts have the same stance label on the same topic, and FALSE if this is not the case. We again use 100 pairs for training, and kept the validation and test sets as-are from the original datasets.

\textbf{Cross versus in-Topic} Five datasets in the benchmark are testing a cross-topic scenario:
argmin, scd, ibmcs, iac, and perspectrum. Two datasets test a within-topic scenario: arc and semeval 2016. We use the topic distributions over train, development, and test sets of these benchmark as-is, see Table~\ref{tab:datasets}.

\subsection{Models} 

\textbf{Architectures} Our bi-encoding models use SETFIT \cite{tunstall2022efficient}, a framework based on Sentence Transformers \cite{reimers2019sentence}. Sentence Transformers is a bi-encoding architecture that encodes more similar sentences as more close in the embedding representations, and less similar sentences as more distant. SETFIT's training approach is contrastive learning  with triplets of [text1, text2, label] on whether the sentences have the same or a different label. The 
sentence encoding is then fed to a classification head, which predicts the class labels. SETFIT was originally designed as a few-shot learning method. 
Our cross-encoding architecture is the RoBERTa model with classification head. 

\textbf{Language Models}
We use all our models as released on the Huggingface hub. Our RoBERTa-large bi-encoding model in the SETFIT framework is sentencebert/RoBERTa-large \cite{reimers2019sentence}. Our NLI model for bi-encoding is SBERT-RoBERTa-NLI  \cite{reimers2019sentence}, trained on Stanford Natural Language Inference dataset \cite{bowman-etal-2015-large} and the Multi-Genre Natural Language Inference Corpus \cite{williams2018broad}. 

Our cross-encoding RoBERTa-large model is RoBERTa-large  \cite{liu2019roberta}. For NLI task knowledge, we use RoBERTa-large-NLI \cite{liu2019roberta} trained with the Multi-Genre Natural Language Inference Corpus \cite{williams2018broad}.

\textbf{Hyperparameters}
For cross-encoding, we optimize our models with Adam with Sparse Categorical Crossentropy. We follow \citet{mosbach2020stability} on stability recommendations for training Transformer models on small datasets, and train for many epochs (20 epochs), have a small batch size (2), and a low learning rate of
$16 \times 10^{-6}$. We save the weights of the epoch with the best loss for development set inference. 
To keep the experiments comparable, we also optimize the bi-encoding models with Adam, train for 20 epochs, a classification head learning rate of  
$16 \times 10^{-6}$ and have a small batch-size (2). SETFIT also requires a step of training the encoding body independent from the classification head. SETFIT learning rate of the contrastive learning while freezing the classification head  \cite{tunstall2022efficient} is $10^{-5}$, and trained for only one epoch as to not over-fit.

\textbf{Software and Hardware}
We use python3.7 with the Huggingface \texttt{Transformers} \cite{wolf2020transformers} library for Transformer models, with Tensorflow back-end. Our code is based on earlier released code by \citet{tunstall2022efficient} and \citet{shnarch2022cluster}. Our experimental code can be found on GitHub.\footnote{\url{https://github.com/myrthereuver/ModelDecisionsStance}} Our cross-encoding experiments were run on two GeForce RTX 2080 Ti GPUs, bi-encoding experiments experiments were run on a single A10 GPU.

\section{Results}

\begin{table*}[t]
\centering
\small{
\begin{tabular}{lllllllcccccc|cccccc}
\toprule

\multicolumn{1}{c}{\textbf{Definition}} &\multicolumn{1}{l}{\textbf{\small{Encoding}}} & \multicolumn{1}{l|}{\textbf{+Task}} & \multicolumn{2}{c|}{\textbf{\textbf{cross-topic }}}  & \multicolumn{3}{l}{\textbf{\textbf{in-topic }}}  \\ 

  \midrule

 && \multicolumn{1}{c}{\textbf{}} & \multicolumn{1}{|c}{\textbf{\textbf{2-class datasets }}} &  \multicolumn{1}{|c|}{\textbf{\textbf{3-class dataset}}} & \multicolumn{2}{|l}{\textbf{\textbf{3+ class datasets}}} \\
  && \multicolumn{1}{c}{\textbf{}} & \multicolumn{1}{|c}{\small{ibmcs, pers, argmin, scd} } &  \multicolumn{1}{|c|}{iac} & \multicolumn{2}{|l}{arc, semeval2016} \\

\midrule

Pro/Con   & \multicolumn{1}{c}{cross-} & \multicolumn{1}{l|}{-} &  \multicolumn{1}{c}{.534 (.052)} &  \multicolumn{1}{c}{.344} &  \multicolumn{1}{c}{.337 (.151)} &&&& \\ 
 
   & \multicolumn{1}{c}{encoding} & \multicolumn{1}{l|}{+NLI} &  \multicolumn{1}{c}{.541 (.035)} & \multicolumn{1}{c}{.345} & \multicolumn{1}{c}{.324 (.123)} \\ 
  
\cmidrule(lr){4-8} 

  & \multicolumn{1}{c}{bi-} & \multicolumn{1}{l|}{+SentSim} & \multicolumn{1}{c}{.554 (.078)}\ & \multicolumn{1}{c}{.280} & .457 (.190) \\

  & \multicolumn{1}{c}{encoding} & \multicolumn{1}{l|}{+NLI}  & \multicolumn{1}{c}{.582 (.038)} & \multicolumn{1}{c}{.288} & .483 (.026) \\ 

\midrule

SSSC &\multicolumn{1}{c}{cross-} & \multicolumn{1}{l|}{-} & \multicolumn{1}{c}{.539 (.120)} & \multicolumn{1}{c}{.389} & .470 (.025)\\ 

   & \multicolumn{1}{c}{encoding} & \multicolumn{1}{l|}{+NLI} & \multicolumn{1}{c}{.623 (.157)} & \multicolumn{1}{c}{.455} & .569 (.032) \\ 
   
\cmidrule(lr){3-6} 

& \multicolumn{1}{c}{bi-} & \multicolumn{1}{l|}{} & \multicolumn{1}{c}{.468 (.036)} & \multicolumn{1}{c}{.525} & .450 (.039)\\

& \multicolumn{1}{c}{encoding} & \multicolumn{1}{l|}{+NLI} & \multicolumn{1}{c}{.543 (.059)} & \multicolumn{1}{c}{.516} & .445 (.020)\\

\bottomrule

\end{tabular}
}
\caption{Average results (in macro F1) over different dataset types (cross/in-topic, 2-class and 3 class datasets) on the few-shot stance experiments.}
\label{tab:hypo1}
\end{table*}

\begin{figure*}
    \centering
\includegraphics[width=\textwidth,scale=.15]{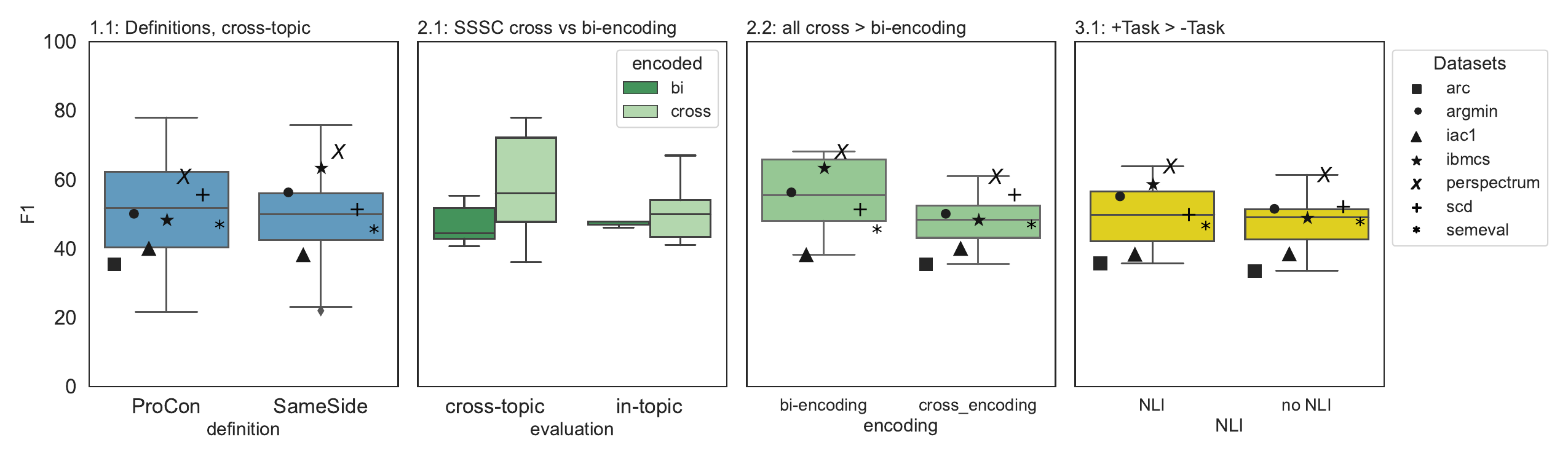}
    \caption{
    Comparisons for Hypothesis 1.1, 2.1, 2.2, and 3.1. Colours of hypotheses are the same as the colours of the decision choices (definition, encoding, and task knowledge) from Figure~\ref{fig:researchprocess}.}
\label{fig:results}
\end{figure*}

\begin{figure}
    \centering
\includegraphics[width=.5\textwidth,scale=.80]{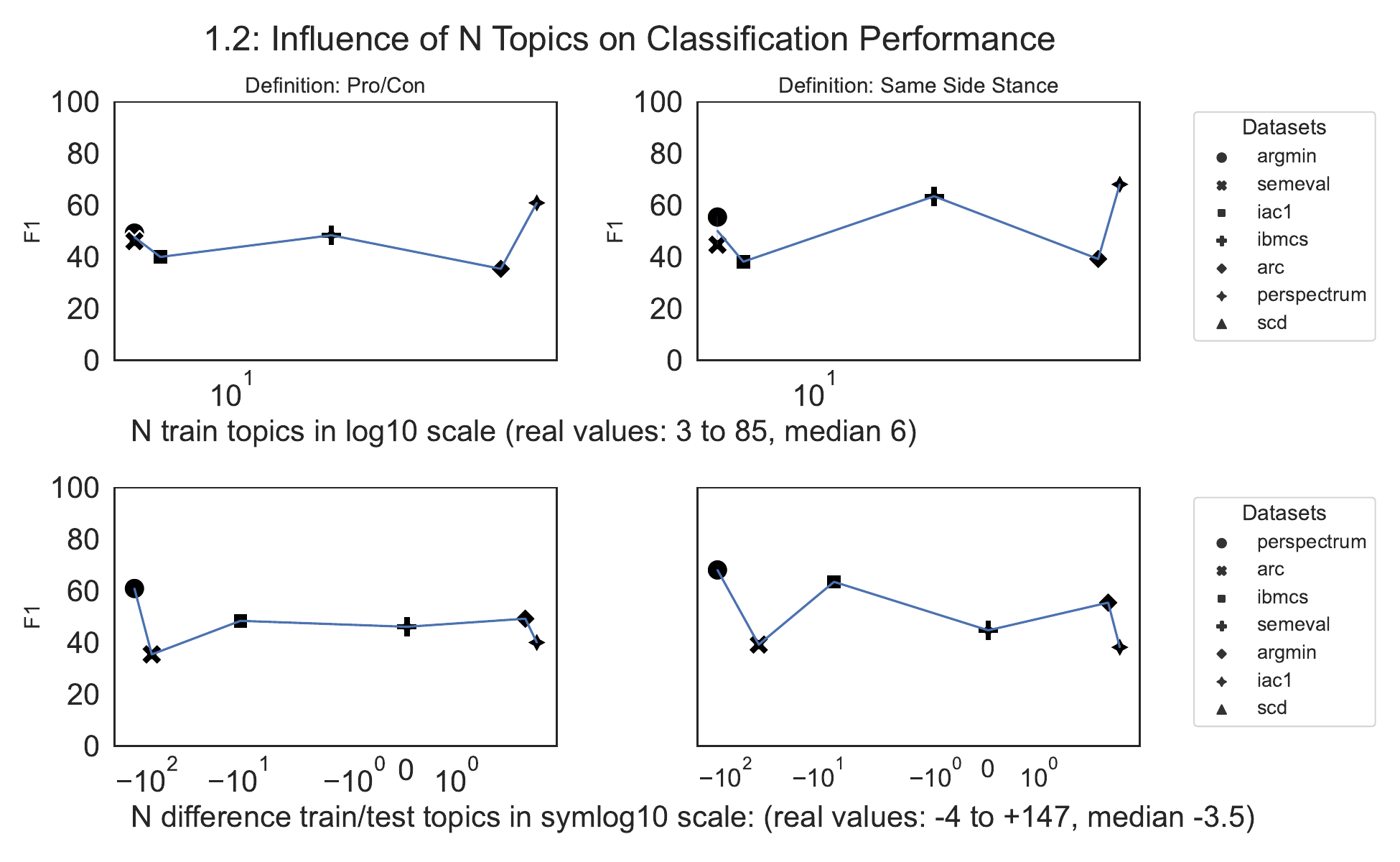}
    \caption{
    Visual representation of the comparisons made for Hypothesis 1.2}
\label{fig:topics}
\end{figure}

Classification results (in macro-F1 performance) over 7 datasets, averaged over modelling choice and dataset specifics such as number of classes, are presented in Table~\ref{tab:hypo1}.
A table with results for the 7 individual datasets is in Appendix~\ref{app:data_result}.
Results concerning Hypotheses 1.1, 2.1, 2.2, and 3.1 are also visually shown in Figure~\ref{fig:results}. Hypothesis 1.2 is separately visualized in Figure~\ref{fig:topics}.

Our results average over 7 stance benchmark datasets and all modelling choices to $F1 =.447, SD = .090)$, with for instance $F1 = .525$ for the \textit{iac} dataset (with SSSC definition, bi-encoding, and without NLI) to $F1 = .766$ for the \textit{perspectrum} dataset (with SSSC definition, cross-encoding and NLI). These are comparable to results in related literature on the same datasets: $F1 =.536$ (\textit{iac}) vs $F1 = .766$ (\textit{perspectrum}) \cite{arakelyan-etal-2023-topic}. This shows our results are sufficiently well-performing for investigating the effects of modelling choices. See Section~\ref{sec:disc} for more in-depth comparisons to earlier work. We interpret results according to our hypotheses in the sections below.

\subsection{Task Definition}
\textbf{Hypothesis 1.1} We expected the SSSC task definition to be more cross-topic robust than the pro/con task definition. Visually, the relevant distinction between task definitions in the cross-topic condition is presented in the first panel of Figure~\ref{fig:results}.

The mean cross-topic performance \textbf{for the Pro/Con definition} over 4 cross-topic datasets in all modelling choices is 
($F1 = .433, SD = .117$). This 
shows less stability than the mean \textbf{cross-topic performance of the SSSC definition} 
($F1 = .507, SD = .035$). SSSC shows higher classification performance ($F1 +.074$), but also a higher standard deviation ($F1 +.082$), meaning more difference across datasets. This seems to confirm our hypothesis of SSSC being more robust, i.e.\ less difference across datasets. However, this is not a fair comparison, as it includes datasets with different numbers of labels. Only 2-label cross-topic datasets show very similar performance in both definitions, with Pro/Con $F1 = .552$, $SD = .051$, SSSC: $F1 = .543$, $SD = .093$. That shows no meaningful difference between task definitions in datasets with the same number of labels.

However, an effect of task definition does appear when considering a relationship with other modelling decisions, or in different individual datasets.  The relationship between SSSC and Pro/Con is different in cross-encoding versus bi-encoding models. In cross-encoding models, SSSC out-performs Pro/Con substantially. This improvement in models with NLI is $F1+ .126$, and $F1+.042$ in F1 without NLI. In bi-encoding models, the SSSC definition shows substantially less improvement: an increase of F1 $+.001$ without NLI, and $+.039$ with NLI. 

\vspace{4mm}
Some individual datasets do show more cross-topic robustness in SSSC, with considerable difference: The first panel of Figure~\ref{fig:results} shows that the two-label \textit{perspectrum} dataset ($X$ symbol) gains almost a .05 in F1 with a SSSC definition (from $F1 = .597$ with Pro/Con to $F1 = .628$ with SSSC stance), while the three-label iac dataset (plotted as black $\triangle$) shows a much larger improvement of > .1 (SSSC: $F1 = .471$, Pro/Con: $F1 = .315$). The \textit{scd} ($+$ symbol) dataset, in some conditions, is less robust with SSSC: $F1 = .461$ for bi-encoding, while $F1 = .436$ for cross-encoding without NLI. With NLI, we see the opposite, and SSSC out-performs the Pro/Con definition. 

 \vspace{4mm}

To summarize: The impact of the task definition leads to mixed results. It is visible in some datasets, but not in all. SSSC does seem to consistently improve results when combined with NLI training, and this is mostly visible in datasets with two Pro/Con stance labels.

\newpage

\textbf{Hypothesis 1.2}
We expected the number of topics in train/test splits to not be connect to performance. We indeed see no effect of the number of topics on classification performance (see Table~\ref{tab:datasets} for the number of topics per dataset). 
The top two subplots in Figure~\ref{fig:topics} represent the relationship between number of training topics and mean classification performance on the different datasets, per task definition. We find no performance increase for datasets with more topics: datasets with best performance are dataset with lowest and highest number of training topics.

The relationship between the number of training and test topics differs per dataset: perspectrum has substantially more test topics than training topics (+147), ibmcs has only slightly more (+8/9), and iac1 (-4) as well as argmin (-3) have less test topics than training topics. The other datasets (arc and SemEval) have no difference in number of topics in train and test sets. The lower two subplots of Figure~\ref{fig:topics} also showed no clear relationship between classification performance and datasets with more, less, or the same number of topics as in test sets.

\subsection{Encoding Choices}\label{sec:arch-hypo}

\textbf{Hypothesis 2.1}
For Same Side Stance, we expected bi-encoding
to have less difference than cross-encoding between in-topic and cross-topic evaluation.
We see this is indeed the case. Figure~\ref{fig:results}, panel 2 shows the difference in cross versus bi-encoding in the Same Side Stance task. Cross-encoding in general out-performs bi-encoding for Same Side Stance, when averaging performance across all other modelling choices.

Bi-encoding SSSC (without NLI) performs in-topic lower (mean $F1 = .450$, SD = .039) than cross-topic (mean $F1 = .469$). The slight difference is $.019$ in F1 between cross and in-topic.
Cross-encoding SSSC (without NLI) shows across in-topic datasets ($F1 = .569$)  slightly higher performance than cross-topic ($F1 = .539$). The difference is $.030$ in F1.

There is a slight difference between cross-topic and within-topic performance for bi-encoding. The difference between in-topic and cross-topic performance is higher for cross-encoding. This confirms our hypothesis that bi-encoding is performing more cross-topic robust, i.e.\ with less difference between in and cross-topic performance. However, different datasets respond differently to encoding choices. Panel 3 of Figure~\ref{fig:results} shows the \textit{IAC1} dataset (plotted as black $\triangle$) improving with cross- over bi-encoding with nearly +.1 F1. Reversely, \textit{ibmcs} (plotted as $\star$) improves with +.1 F1 with bi-encoding over cross-encoding.

\textbf{Hypothesis 2.2}
Our hypothesis was that cross-encoding would perform better than bi-encoding in stance classification, following previous literature. Overall, we found the oppposite, with bi-encoding showing slightly higher classification performance than cross-encoding. However, we found a multi-faceted relationship, with other modelling choices influencing performance, and a strong difference between different datasets. Same Side Stance responds better to cross-encoding ($F1 = .540$, $SD = .120$) than bi-encoding ($F1 = .491$,  $SD = .054$), while Pro/Con performs similarly with bi-encoding ($F1 = .451$,  $SD = .121$) and cross-encoding ($F1 = 460$, $SD = .151$). 

\subsection{Task Knowledge}

\textbf{Hypothesis 3.1}
We hypothesized that models fine-tuned on an auxiliary task would outperform models without additional fine-tuning. Results of our experiments with NLI show that NLI indeed mostly improves classification performance, but not always. Overall, +NLI ($F1 = .502$) does not clearly outperform the modelling choice without NLI ($F1 = .469$), mostly due to the lack of improvement in in-topic datasets. However, looking at other modelling choices, the positive effect of +NLI is very clear. In the Pro/Con definition we in general see that NLI helps, but not as much as for SSSC, especially in the cross-encoding models. With the Same Side Stance definition, cross-encoding models show an average increase of $F1 = .084$ in 2-class datasets, which is substantial. Similarly, Same Side Stance models with bi-encoding + NLI show a considerable and large improvement for some datasets (perspectrum, scd, arc, sem2016) compared to bi-encoding without NLI. Some of our cross-encoding NLI models perform much higher than the reported mean, up to $F1 = .766$ for \textit{perspectrum} and $F1 = .734$ for \textit{ibmcs}.

\section{Discussion}\label{sec:disc}

\textbf{Results related to hypotheses} Some of our hypotheses and the claims in previous literature \cite{stein2021same} were confirmed by our results, but only in some conditions: Same Side Stance is more cross-topic robust, but considering different datasets and other modelling choices shows that the result is different for different encoding choices. We also saw no performance increase with more topics in training, which we expected.

For encoding, we found an influence of other modelling choices: we expected better performance with cross-encoding overall, and a bi-encoding to aid cross-topic robustness in Same Side Stance. We found that in fact cross-encoding out-performs bi-encoding in same-side stance, but some datasets performing better with cross-encoding, while others with bi-encoding.

Adding NLI as auxiliary knowledge improved the classification performance of our models substantially, which becomes visible when looking at the effect on other modelling choices. \citet{heinisch2022overview} and \citet{van2022will} already found that NLI knowledge aids performance in argument-related tasks. On the other hand, it is surprising that NLI does not improve classification performance for all datasets. Some datasets respond exceptionally well to adding knowledge from NLI, while other datasets do not show this improvement. 
The variance in performance across datasets makes it difficult to choose one few-shot modelling approach for stance on all datasets. 

\textbf{Comparison to results in related work}  
\citet{varadarajan-etal-2022-detecting} conclude that a cross-encoder Transformer can generalize Same Side Stance to previously unseen topics with only a few training examples, and can do so with similar performance to in the in-topic setting. Their RoBERTa model obtained $F1 = .718$ on SSSC with three classes (similar, different, unrelated), 
when trained on 32 unrelated topics and tested on 5 unseen topics.
 However, earlier computational argumentation research has shown that having an `unrelated' or `no argument' class can inflate performance \cite{reuver2021stance}. Our average performance of the Same Side Stance task with cross-encoding architecture over four different cross-topic stance datasets with two classes is $F1=.557$ (training with 100 examples), but with a large deviation over different datasets. The model trained on the \textit{perspectrum} dataset (80 train topics, 30 other test topics) performs well with $F1 = .702$, with adding NLI knowledge boosting this performance to $F1 = .776$. Other datasets, e.g.\ the \textit{argmin} dataset (five training topics, one other test topic), perform much below this with $F1 = .411$, but also with a substantial improvement with NLI to $F1 =.564$. Reporting one average results over one dataset appears insufficient to show robust effects of task definitions such as Same Side Stance. 
 
 \citet{korner2021classifying} perform full dataset learning on one Same Side Stance dataset (60,362 train pairs, 1 train topic, testing on 32,842 test examples of another topic), and achieved a highest performance of $F1 = .624$ when carefully considering a clean train/test-split, while an earlier shared task evaluation \cite{ollinger2020same} on a version of this dataset with more train/test overlap \cite{stein2021same} led to a highest F1 of $.737$. These cross-topic results with models trained on a full dataset are comparable to or even worse than some of our cross-topic \textit{few-shot} NLI models on several of our datasets: \textit{ibmcs} (21 training, 30 test topics) with $F1 = .785$, and \textit{perspectrum} (80 train topics, 30 other test topics) with $F1 = .772$. One possible explanation is that topic diversity in the training set allows for better generalization, which is what \citet{varadarajan-etal-2022-detecting} found. However, we found no clear relationship between topic diversity and classification performance.

 \citet{arakelyan-etal-2023-topic} focus on optimizing data selection with topic diversity rather than optimizing modelling decisions. They select < 10\% of each stance dataset. Comparing our best results per dataset to theirs shows comparable results. \textit{Perspectrum} receives $F1 = .83$ with their optimized sample and .77 with a random sample, which is comparable to our best result for that dataset ($F1 = .766$). For the \textit{iac} dataset they obtain $F1 = .569$, and $F1 =.42$ with random sampling. This is comparable to our result with the best modelling choices for that dataset ($F1 = .536$). This shows that in few-shot learning, optimizing for modelling decisions can have a similar performance improvement to optimizing data subsets. Recently, \citet{waldis-etal-2024-dive} investigated differently pre-trained models, which is a modelling choice we did not explore. They find diverse pre-training objectives allow for better cross-topic stance capabilities.
 
\textbf{Dataset characteristics} 
Different modelling choices have different effects on different stance datasets. The Pro/Con datasets that perform best are the \textit{perspectrum} dataset ($F1 = .766$ with NLI) and the \textit{ibmcs} dataset ($F1 =.734$ with NLI), while several other datasets with the exact same modelling choices (Pro/Con with NLI) perform substantially lower, such as the \textit{scd} dataset ($F1 =.428$) and \textit{argmin} ($F1 = .564$). All of these are datasets with 2 Pro/Con stance labels, so the difference cannot be due to a different number of labels. Their topic contents are slightly different: the most successful datasets are the ones where the training data size has relatively many topics: 21-22\footnote{The number of topics differs with +/- one to three topics per task definition, see Table~\ref{tab:datasets}} (\textit{ibmcs}) and 80-82 (\textit{perspectrum}), while the least successful datasets have fewer topics in the training data (five for \textit{argmin} and three for \textit{scd}). However, a clear linear relationship between number of topics in training and performance, or even training-test difference, was not visible in Figure~\ref{fig:topics} when plotting average performance of all models with the same dataset. 
Table~\ref{tab:cross-topic-2} in Appendix~\ref{app:data_result} does seem to show that datasets with more topics perform better, but mostly with NLI models.

It appears that stance datasets with the highest performances contain texts from websites specifically aimed at debating. \textit{Perspectrum} \cite{chen-etal-2019-seeing} contains texts from debate websites and its topics are formal proposals, such as ``The European Union should adopt a single working language through which to operate". The \textit{ibmcs} dataset \cite{bar-haim-etal-2017-stance} has similar content. 
In contrast, \textit{argmin} \cite{stab-etal-2018-cross} is purposely built from heterogeneous web documents, and contains one-word topics such as ``abortion". 

\citet{ng2022my} looked into differences of seven stance datasets. One of their findings was that in different 80/20\% train/test splits, performance fluctuates substantially. This fits our results: we found performance inconsistency in different datasets, while they found performance inconsistency \textit{within} stance datasets.

\section{Conclusion}
This paper compares different modelling choices for few-shot modelling of stance for the goal of diversifying news.
We investigate two task definitions (Same Side Stance and Pro/Con), two architectures (cross and bi-encoding), and the addition of related task knowledge (NLI) with five pre-registered hypotheses on seven stance benchmark datasets. The effect of the Same Side Stance definition on performance differs per dataset and other modelling choice, and we found no clear relationship between number of training topics and performance. Cross-encoding generally out-performs bi-encoding, but some datasets show the opposite effect. Adding NLI training to our models gives considerable improvement for some datasets, but inconsistent results for others.
We conclude that experiments beyond one dataset or modelling choice are essential for finding robust effects of modelling the concept `stance'.  A viewpoint-diverse news recommender can use robust cross-topic stance models to measure stance differences between news articles, in order to recommend new viewpoints to newsreaders.

\newpage
\section*{Limitations}
\paragraph{Dataset Limitations} Our results are limited in several ways. First of all, these experimental results only reflect outcomes on English-language datasets \cite{benderrule}, and then mostly datasets centered on socio-political debates from the English-speaking United States, such as gun control, abortion, and the US-Iraq war. This limits especially the cross-topic experiments, which would benefit from more topic diversity.

\paragraph{Computational limitations } The models presented in this work are dependent on access to GPU computing clusters. This will for reproducibility imply dependence on (third-party) cloud computing systems. We would like to state that unequal access to computing resources, and an non-transparent explanation of computing resources used in NLP, is an a prominent problem in recent GPU resource-hungry NLP work. See \citet{lee2023surveying} for a first quantified exploration of the effects of such inequality on Natural Language Processing research and the research community.

\paragraph{Complexity of opinions} This paper represents human viewpoints as stances in favor or against debate topics. This is a simplification of real human opinion and human debate. Opinions are complex, subject to change, and contextual to specific topics and contexts \cite{joseph-etal-2021-mis}. Reducing this to a stance label that never changes would make this unfair and also not representative of humans in society. We recommend stakeholders to be careful when attempting to represent humans, human reading behaviours, or any other aspect with a stance label that is permanent, and to carefully consider context factors when doing so.

\section*{Ethical Considerations}

\paragraph{Preregistration and responsible science}
This paper describes a hypothesis-driven study, as we pre-registered our hypotheses before we started our experiments. Preregistration is relatively new to NLP, first introduced to our community with extensive recommendations by \citet{van2021preregistering}, and discussed by \citet{sogaard-etal-2023-two}. Our experiments used preregistration with several aims related to ethical and responsible science. Firstly, we wanted to be transparent and precise about our research contributions, which pre-registered hypotheses allow us to do. Secondly, pre-registering our experiments allows us to be focused and limited in the amount of compute needed to answer our research questions. This has benefits for a large problem in current experimental NLP work: large pre-trained models have a large the climate impact \cite{hershcovich-etal-2022-towards}. Another benefit of this controlled use of compute is for research groups having less access to computational resources, who can with preregistration clearly target specific hypotheses rather than running larger experiments aimed at finding performance increases.
Preregistration also encourages researchers to not only present positive results, but also mixed results. NLP has a publishing culture with difficulty of publishing negative results \cite{parra-escartin-etal-2017-ethical}, and a hard time of publishing research that has more negative results than earlier published positive results. We therefore consider preregistration a potential tool for more responsible and ethical publishing practices, and recommend other authors to explore it.

\paragraph{Detecting difference in opinion}
Our study is intended as a building block towards building computational models that are more able to detect a diversity of opinions, and our ultimate goal is to design an opinion-diverse news information platform. This is a goal with clear normative implications: we aim to promote diversity and democratic, deliberative debate. However, work on computational argumentation is a double-edged sword, as models have a dual use: detection of different arguments can also be used to detect and then \textit{suppress} opinion diversity, rather than promote diversity. Another possible negative use case is one to detect opinionated expressions that are against certain governments or policies. The latter use-case could lead to the censorship of minorities or dissenting opinions by governments, companies, or other powerful actors. We explicitly condemn any of such uses of our models.

\paragraph{}

\section*{Acknowledgements}
This research is part of the \textit{Rethinking News Algorithms} project (2020-2024), funded through the Open Competition Digitalization Humanities and Social Science (grant nr 406.D1.19.073) by the Netherlands Organization of Scientific Research (NWO). Our computing was done through SURF Research Cloud, a national supercomputer infrastructure in the Netherlands also funded by the NWO. Thanks to Urja Khurana, Michiel van der Meer, and other PhDs from CLTL for their helpful feedback. We would also like to thank all anonymous reviewers, whose comments improved both this version and earlier versions of this paper. All remaining errors or unclarities are our own.

\newpage
\section{Bibliographical References}\label{sec:reference}

\bibliographystyle{lrec-coling2024-natbib}
\bibliography{lrec-coling2024-example}

\begin{thebibliography}{7}
\expandafter\ifx\csname natexlab\endcsname\relax\def\natexlab#1{#1}\fi

\bibitem[{Bar-Haim et~al.(2017)Bar-Haim, Bhattacharya, Dinuzzo, Saha, and Slonim}]{data4}
Bar-Haim, Roy and Bhattacharya, Indrajit and Dinuzzo, Francesco and Saha, Amrita and Slonim, Noam. 2017.
\newblock \href {https://aclanthology.org/E17-1024} {\emph{Stance Classification of Context-Dependent Claims}}.
\newblock Association for Computational Linguistics.

\bibitem[{Chen et~al.(2019)Chen, Khashabi, Yin, Callison-Burch, and Roth}]{data5}
Chen, Sihao and Khashabi, Daniel and Yin, Wenpeng and Callison-Burch, Chris and Roth, Dan. 2019.
\newblock \emph{Seeing Things from a Different Angle:Discovering Diverse Perspectives about Claims}.
\newblock Association for Computational Linguistics.
\newblock PID \href{10.18653/v1/N19-1053}{10.18653/v1/N19-1053}.

\bibitem[{Habernal et~al.(2018)Habernal, Wachsmuth, Gurevych, and Stein}]{arc}
Habernal, Ivan and Wachsmuth, Henning and Gurevych, Iryna and Stein, Benno. 2018.
\newblock \emph{The Argument Reasoning Comprehension Task: Identification and Reconstruction of Implicit Warrants}.
\newblock Proceedings of NAACL-HLT 2018, 1.0, ISLRN \href{https://www.islrn.org/resources/10.18653/v1/N18-1175}{10.18653/v1/N18-1175}.

\bibitem[{Hasan and Ng(2013)}]{data3}
Hasan, Kazi Saidul and Ng, Vincent. 2013.
\newblock \href {https://aclanthology.org/I13-1191} {\emph{Stance Classification of Ideological Debates: Data, Models, Features, and Constraints}}.
\newblock Asian Federation of Natural Language Processing.

\bibitem[{Mohammad et~al.(2016)Mohammad, Kiritchenko, Sobhani, Zhu, and Cherry}]{mohammad2016semeval}
Mohammad, Saif and Kiritchenko, Svetlana and Sobhani, Parinaz and Zhu, Xiaodan and Cherry, Colin. 2016.
\newblock \emph{Semeval-2016 task 6: Detecting stance in tweets}.

\bibitem[{Stab et~al.(2018)Stab, Miller, Schiller, Rai, and Gurevych}]{argmin}
Stab, Christian and Miller, Tristan and Schiller, Benjamin and Rai, Pranav and Gurevych, Iryna. 2018.
\newblock \href {https://aclanthology.org/D18-1402} {\emph{Cross-topic Argument Mining from Heterogeneous Sources}}.
\newblock Association for Computational Linguistics.
\newblock PID \href{10.18653/v1/D18-1402}{10.18653/v1/D18-1402}.

\bibitem[{Walker et~al.(2012)Walker, Tree, Anand, Abbott, and King}]{walker}
Walker, Marilyn and Tree, Jean Fox and Anand, Pranav and Abbott, Rob and King, Joseph. 2012.
\newblock \emph{A Corpus for Research on Deliberation and Debate}.
\newblock European Language Resources Association (ELRA), ISLRN \href{https://www.islrn.org/resources/http://www.lrec-conf.org/proceedings/lrec2012/pdf/1078\_Paper.pdf}{http://www.lrec-conf.org/proceedings/lrec2012/pdf/1078\_Paper.pdf}.

\end{thebibliography}


\begin{thebibliography}{55}
\expandafter\ifx\csname natexlab\endcsname\relax\def\natexlab#1{#1}\fi

\bibitem[{Ajjour et~al.(2023)Ajjour, Kiesel, Stein, and Potthast}]{ajjour2023topic}
Yamen Ajjour, Johannes Kiesel, Benno Stein, and Martin Potthast. 2023.
\newblock \href {https://doi.org/10.18653/v1/2023.findings-eacl.104} {Topic ontologies for arguments}.
\newblock In \emph{Findings of the Association for Computational Linguistics: EACL 2023}, pages 1411--1427, Dubrovnik, Croatia. Association for Computational Linguistics.

\bibitem[{Alam et~al.(2022)Alam, Iana, Grote, Ludwig, M{\"u}ller, and Paulheim}]{alam2022towards}
Mehwish Alam, Andreea Iana, Alexander Grote, Katharina Ludwig, Philipp M{\"u}ller, and Heiko Paulheim. 2022.
\newblock Towards analyzing the bias of news recommender systems using sentiment and stance detection.
\newblock In \emph{Companion Proceedings of the Web Conference 2022}, pages 448--457.

\bibitem[{Albalak et~al.(2022)Albalak, Tuan, Jandaghi, Pryor, Yoffe, Ramachandran, Getoor, Pujara, and Wang}]{Albalak2022FETAAB}
Alon Albalak, Yi-Lin Tuan, Pegah Jandaghi, Connor Pryor, Luke Yoffe, Deepak Ramachandran, Lise Getoor, Jay Pujara, and William~Yang Wang. 2022.
\newblock Feta: A benchmark for few-sample task transfer in open-domain dialogue.

\bibitem[{Allaway et~al.(2021)Allaway, Srikanth, and Mckeown}]{allaway2021adversarial}
Emily Allaway, Malavika Srikanth, and Kathleen Mckeown. 2021.
\newblock Adversarial learning for zero-shot stance detection on social media.
\newblock In \emph{Proceedings of the 2021 Conference of the North American Chapter of the Association for Computational Linguistics: Human Language Technologies}, pages 4756--4767.

\bibitem[{Alshomary and Wachsmuth(2019)}]{alshomary2021siamese}
Milad Alshomary and Henning Wachsmuth. 2019.
\newblock Siamese neural network for same side stance classification.
\newblock In \emph{{Proceedings of the Same Side Stance Classification Shared Task organized as a part of the 6th Workshop on Argument Mining (ArgMining 2019)}}.

\bibitem[{Arakelyan et~al.(2023)Arakelyan, Arora, and Augenstein}]{arakelyan-etal-2023-topic}
Erik Arakelyan, Arnav Arora, and Isabelle Augenstein. 2023.
\newblock \href {https://doi.org/10.18653/v1/2023.acl-long.752} {Topic-guided sampling for data-efficient multi-domain stance detection}.
\newblock In \emph{Proceedings of the 61st Annual Meeting of the Association for Computational Linguistics (Volume 1: Long Papers)}, pages 13448--13464, Toronto, Canada. Association for Computational Linguistics.

\bibitem[{Bar-Haim et~al.(2017)Bar-Haim, Bhattacharya, Dinuzzo, Saha, and Slonim}]{bar-haim-etal-2017-stance}
Roy Bar-Haim, Indrajit Bhattacharya, Francesco Dinuzzo, Amrita Saha, and Noam Slonim. 2017.
\newblock \href {https://aclanthology.org/E17-1024} {Stance classification of context-dependent claims}.
\newblock In \emph{Proceedings of the 15th Conference of the {E}uropean Chapter of the Association for Computational Linguistics: Volume 1, Long Papers}, pages 251--261, Valencia, Spain. Association for Computational Linguistics.

\bibitem[{Beck et~al.(2023)Beck, Waldis, and Gurevych}]{beck2023robust}
Tilman Beck, Andreas Waldis, and Iryna Gurevych. 2023.
\newblock Robust integration of contextual information for cross-target stance detection.
\newblock In \emph{Proceedings of the The 12th Joint Conference on Lexical and Computational Semantics (* SEM 2023)}, pages 494--511.

\bibitem[{Bender(2019)}]{benderrule}
Emily Bender. 2019.
\newblock The\# benderrule: On naming the languages we study and why it matters.
\newblock \emph{The Gradient}, 14:34.

\bibitem[{Bowman et~al.(2015)Bowman, Angeli, Potts, and Manning}]{bowman-etal-2015-large}
Samuel~R. Bowman, Gabor Angeli, Christopher Potts, and Christopher~D. Manning. 2015.
\newblock \href {https://doi.org/10.18653/v1/D15-1075} {A large annotated corpus for learning natural language inference}.
\newblock In \emph{Proceedings of the 2015 Conference on Empirical Methods in Natural Language Processing}, pages 632--642, Lisbon, Portugal. Association for Computational Linguistics.

\bibitem[{Chen et~al.(2019)Chen, Khashabi, Yin, Callison-Burch, and Roth}]{chen-etal-2019-seeing}
Sihao Chen, Daniel Khashabi, Wenpeng Yin, Chris Callison-Burch, and Dan Roth. 2019.
\newblock \href {https://doi.org/10.18653/v1/N19-1053} {Seeing things from a different angle:discovering diverse perspectives about claims}.
\newblock In \emph{Proceedings of the 2019 Conference of the North {A}merican Chapter of the Association for Computational Linguistics: Human Language Technologies, Volume 1 (Long and Short Papers)}, pages 542--557, Minneapolis, Minnesota. Association for Computational Linguistics.

\bibitem[{Du~Bois(2007)}]{du2007stance}
John~W Du~Bois. 2007.
\newblock The stance triangle.
\newblock \emph{Stancetaking in discourse: Subjectivity, evaluation, interaction}, 164(3):139--182.

\bibitem[{Figueras et~al.(2023)Figueras, Baucells, and Caselli}]{figueras2023dynamic}
Blanca Figueras, Irene Baucells, and Tommaso Caselli. 2023.
\newblock Dynamic stance: Modeling discussions by labeling the interactions.
\newblock In \emph{Findings of the Association for Computational Linguistics: EMNLP 2023}, pages 6503--6515.

\bibitem[{Hardalov et~al.(2021)Hardalov, Arora, Nakov, and Augenstein}]{hardalov-etal-2021-cross}
Momchil Hardalov, Arnav Arora, Preslav Nakov, and Isabelle Augenstein. 2021.
\newblock \href {https://doi.org/10.18653/v1/2021.emnlp-main.710} {Cross-domain label-adaptive stance detection}.
\newblock In \emph{Proceedings of the 2021 Conference on Empirical Methods in Natural Language Processing}, pages 9011--9028, Online and Punta Cana, Dominican Republic. Association for Computational Linguistics.

\bibitem[{Heinisch et~al.(2022)Heinisch, Frank, Opitz, Plenz, and Cimiano}]{heinisch2022overview}
Philipp Heinisch, Anette Frank, Juri Opitz, Moritz Plenz, and Philipp Cimiano. 2022.
\newblock Overview of the 2022 validity and novelty prediction shared task.
\newblock In \emph{Proceedings of the 9th Workshop on Argument Mining}, pages 84--94.

\bibitem[{Helberger(2019)}]{helberger2019democratic}
Natali Helberger. 2019.
\newblock On the democratic role of news recommenders.
\newblock \emph{Digital Journalism}, 7(8):993--1012.

\bibitem[{Hershcovich et~al.(2022)Hershcovich, Webersinke, Kraus, Bingler, and Leippold}]{hershcovich-etal-2022-towards}
Daniel Hershcovich, Nicolas Webersinke, Mathias Kraus, Julia Bingler, and Markus Leippold. 2022.
\newblock \href {https://doi.org/10.18653/v1/2022.emnlp-main.159} {Towards climate awareness in {NLP} research}.
\newblock In \emph{Proceedings of the 2022 Conference on Empirical Methods in Natural Language Processing}, pages 2480--2494, Abu Dhabi, United Arab Emirates. Association for Computational Linguistics.

\bibitem[{Hossain et~al.(2020)Hossain, Logan~IV, Ugarte, Matsubara, Young, and Singh}]{hossain2020covidlies}
Tamanna Hossain, Robert~L. Logan~IV, Arjuna Ugarte, Yoshitomo Matsubara, Sean Young, and Sameer Singh. 2020.
\newblock \href {https://doi.org/10.18653/v1/2020.nlpcovid19-2.11} {{COVIDL}ies: Detecting {COVID}-19 misinformation on social media}.
\newblock In \emph{Proceedings of the 1st Workshop on {NLP} for {COVID}-19 (Part 2) at {EMNLP} 2020}, Online. Association for Computational Linguistics.

\bibitem[{Hou et~al.(2022)Hou, van~der Putten, and Verberne}]{hou2022covmis}
Yanfang Hou, Peter van~der Putten, and Suzan Verberne. 2022.
\newblock \emph{The COVMis-Stance dataset: Stance Detection on Twitter for COVID-19 Misinformation}.

\bibitem[{Jakobsen et~al.(2021)Jakobsen, Barrett, and S{\o}gaard}]{jakobsen2021spurious}
Terne Sasha~Thorn Jakobsen, Maria Barrett, and Anders S{\o}gaard. 2021.
\newblock Spurious correlations in cross-topic argument mining.
\newblock In \emph{Proceedings of* SEM 2021: The Tenth Joint Conference on Lexical and Computational Semantics}, pages 263--277.

\bibitem[{Joseph et~al.(2021)Joseph, Shugars, Gallagher, Green, Quintana~Math{\'e}, An, and Lazer}]{joseph-etal-2021-mis}
Kenneth Joseph, Sarah Shugars, Ryan Gallagher, Jon Green, Alexi Quintana~Math{\'e}, Zijian An, and David Lazer. 2021.
\newblock \href {https://doi.org/10.18653/v1/2021.emnlp-main.27} {(mis)alignment between stance expressed in social media data and public opinion surveys}.
\newblock In \emph{Proceedings of the 2021 Conference on Empirical Methods in Natural Language Processing}, pages 312--324, Online and Punta Cana, Dominican Republic. Association for Computational Linguistics.

\bibitem[{K{\"o}rner et~al.(2021)K{\"o}rner, Wiedemann, Hakimi, Heyer, and Potthast}]{korner2021classifying}
Erik K{\"o}rner, Gregor Wiedemann, Ahmad~Dawar Hakimi, Gerhard Heyer, and Martin Potthast. 2021.
\newblock On classifying whether two texts are on the same side of an argument.
\newblock In \emph{Proceedings of the 2021 conference on empirical methods in natural language processing}, pages 10130--10138.

\bibitem[{K{\"u}{\c{c}}{\"u}k and Can(2020)}]{kuccuk2020stance}
Dilek K{\"u}{\c{c}}{\"u}k and Fazli Can. 2020.
\newblock Stance detection: A survey.
\newblock \emph{ACM Computing Surveys (CSUR)}, 53(1):1--37.

\bibitem[{Lee et~al.(2023)Lee, Puerto, van Aken, Arase, Forde, Derczynski, R{\"u}ckl{\'e}, Gurevych, Schwartz, Strubell et~al.}]{lee2023surveying}
Ji-Ung Lee, Haritz Puerto, Betty van Aken, Yuki Arase, Jessica~Zosa Forde, Leon Derczynski, Andreas R{\"u}ckl{\'e}, Iryna Gurevych, Roy Schwartz, Emma Strubell, et~al. 2023.
\newblock Surveying (dis) parities and concerns of compute hungry nlp research.
\newblock \emph{arXiv preprint arXiv:2306.16900}.

\bibitem[{Liu et~al.(2021)Liu, Lin, Tan, and Wang}]{liu2021enhancing}
Rui Liu, Zheng Lin, Yutong Tan, and Weiping Wang. 2021.
\newblock Enhancing zero-shot and few-shot stance detection with commonsense knowledge graph.
\newblock In \emph{Findings of the Association for Computational Linguistics: ACL-IJCNLP 2021}, pages 3152--3157.

\bibitem[{Liu et~al.(2019)Liu, Ott, Goyal, Du, Joshi, Chen, Levy, Lewis, Zettlemoyer, and Stoyanov}]{liu2019roberta}
Yinhan Liu, Myle Ott, Naman Goyal, Jingfei Du, Mandar Joshi, Danqi Chen, Omer Levy, Mike Lewis, Luke Zettlemoyer, and Veselin Stoyanov. 2019.
\newblock Roberta: A robustly optimized bert pretraining approach.
\newblock \emph{arXiv preprint arXiv:1907.11692}.

\bibitem[{Mattis et~al.(2022)Mattis, Masur, M{\"o}ller, and van Atteveldt}]{mattis2022nudging}
Nicolas Mattis, Philipp Masur, Judith M{\"o}ller, and Wouter van Atteveldt. 2022.
\newblock \emph{Nudging towards news diversity: A theoretical framework for facilitating diverse news consumption through recommender design}.
\newblock SAGE Publications Sage UK: London, England.

\bibitem[{Mosbach et~al.(2020)Mosbach, Andriushchenko, and Klakow}]{mosbach2020stability}
Marius Mosbach, Maksym Andriushchenko, and Dietrich Klakow. 2020.
\newblock \emph{On the stability of fine-tuning bert: Misconceptions, explanations, and strong baselines}.

\bibitem[{Mulder et~al.(2021)Mulder, Inel, Oosterman, and Tintarev}]{mulder2021operationalizing}
Mats Mulder, Oana Inel, Jasper Oosterman, and Nava Tintarev. 2021.
\newblock Operationalizing framing to support multiperspective recommendations of opinion pieces.
\newblock In \emph{Proceedings of the 2021 ACM conference on fairness, accountability, and transparency}, pages 478--488.

\bibitem[{Ng and Carley(2022)}]{ng2022my}
Lynnette Hui~Xian Ng and Kathleen~M Carley. 2022.
\newblock Is my stance the same as your stance? a cross validation study of stance detection datasets.
\newblock \emph{Information Processing \& Management}, 59(6):103070.

\bibitem[{Nie et~al.(2020)Nie, Williams, Dinan, Bansal, Weston, and Kiela}]{nie-etal-2020-adversarial}
Yixin Nie, Adina Williams, Emily Dinan, Mohit Bansal, Jason Weston, and Douwe Kiela. 2020.
\newblock \href {https://doi.org/10.18653/v1/2020.acl-main.441} {Adversarial {NLI}: A new benchmark for natural language understanding}.
\newblock In \emph{Proceedings of the 58th Annual Meeting of the Association for Computational Linguistics}, pages 4885--4901, Online. Association for Computational Linguistics.

\bibitem[{Ollinger et~al.(2020)Ollinger, Dumani, Sahitaj, Bergmann, and Schenkel}]{ollinger2020same}
Stefan Ollinger, Lorik Dumani, Premtim Sahitaj, Ralph Bergmann, and Ralf Schenkel. 2020.
\newblock Same side stance classification task: Facilitating argument stance classification by fine-tuning a bert model.
\newblock \emph{arXiv preprint arXiv:2004.11163}.

\bibitem[{Parra~Escart{\'\i}n et~al.(2017)Parra~Escart{\'\i}n, Reijers, Lynn, Moorkens, Way, and Liu}]{parra-escartin-etal-2017-ethical}
Carla Parra~Escart{\'\i}n, Wessel Reijers, Teresa Lynn, Joss Moorkens, Andy Way, and Chao-Hong Liu. 2017.
\newblock \href {https://doi.org/10.18653/v1/W17-1608} {Ethical considerations in {NLP} shared tasks}.
\newblock In \emph{Proceedings of the First {ACL} Workshop on Ethics in Natural Language Processing}, pages 66--73, Valencia, Spain. Association for Computational Linguistics.

\bibitem[{Popat et~al.(2019)Popat, Mukherjee, Yates, and Weikum}]{popat2019stancy}
Kashyap Popat, Subhabrata Mukherjee, Andrew Yates, and Gerhard Weikum. 2019.
\newblock Stancy: Stance classification based on consistency cues.
\newblock In \emph{Proceedings of the 2019 Conference on Empirical Methods in Natural Language Processing and the 9th International Joint Conference on Natural Language Processing (EMNLP-IJCNLP)}, pages 6413--6418.

\bibitem[{Pruksachatkun et~al.(2020)Pruksachatkun, Phang, Liu, Htut, Zhang, Pang, Vania, Kann, and Bowman}]{pruksachatkun2020intermediate}
Yada Pruksachatkun, Jason Phang, Haokun Liu, Phu~Mon Htut, Xiaoyi Zhang, Richard~Yuanzhe Pang, Clara Vania, Katharina Kann, and Samuel Bowman. 2020.
\newblock Intermediate-task transfer learning with pretrained language models: When and why does it work?
\newblock In \emph{Proceedings of the 58th Annual Meeting of the Association for Computational Linguistics}, pages 5231--5247.

\bibitem[{Reimers and Gurevych(2019)}]{reimers2019sentence}
Nils Reimers and Iryna Gurevych. 2019.
\newblock Sentence-bert: Sentence embeddings using siamese bert-networks.
\newblock In \emph{Proceedings of the 2019 Conference on Empirical Methods in Natural Language Processing and the 9th International Joint Conference on Natural Language Processing (EMNLP-IJCNLP)}, pages 3982--3992.

\bibitem[{Reimers et~al.(2019)Reimers, Schiller, Beck, Daxenberger, Stab, and Gurevych}]{reimers2019classification}
Nils Reimers, Benjamin Schiller, Tilman Beck, Johannes Daxenberger, Christian Stab, and Iryna Gurevych. 2019.
\newblock Classification and clustering of arguments with contextualized word embeddings.
\newblock In \emph{Proceedings of the 57th Annual Meeting of the Association for Computational Linguistics}, pages 567--578.

\bibitem[{Reuver et~al.(2021)Reuver, Verberne, Morante, and Fokkens}]{reuver2021stance}
Myrthe Reuver, Suzan Verberne, Roser Morante, and Antske Fokkens. 2021.
\newblock Is stance detection topic-independent and cross-topic generalizable?-a reproduction study.
\newblock In \emph{Proceedings of the 8th Workshop on Argument Mining}, pages 46--56.

\bibitem[{Schiller et~al.(2021)Schiller, Daxenberger, and Gurevych}]{schiller2021stance}
Benjamin Schiller, Johannes Daxenberger, and Iryna Gurevych. 2021.
\newblock Stance detection benchmark: How robust is your stance detection?
\newblock \emph{KI-K{\"u}nstliche Intelligenz}, pages 1--13.

\bibitem[{Shnarch et~al.(2022)Shnarch, Gera, Halfon, Dankin, Choshen, Aharonov, and Slonim}]{shnarch2022cluster}
Eyal Shnarch, Ariel Gera, Alon Halfon, Lena Dankin, Leshem Choshen, Ranit Aharonov, and Noam Slonim. 2022.
\newblock Cluster \& tune: Boost cold start performance in text classification.
\newblock In \emph{Proceedings of the 60th Annual Meeting of the Association for Computational Linguistics (Volume 1: Long Papers)}, pages 7639--7653.

\bibitem[{S{\o}gaard et~al.(2023)S{\o}gaard, Hershcovich, and de~Lhoneux}]{sogaard-etal-2023-two}
Anders S{\o}gaard, Daniel Hershcovich, and Miryam de~Lhoneux. 2023.
\newblock \href {https://doi.org/10.18653/v1/2023.eacl-main.6} {A two-sided discussion of preregistration of {NLP} research}.
\newblock In \emph{Proceedings of the 17th Conference of the European Chapter of the Association for Computational Linguistics}, pages 83--93, Dubrovnik, Croatia. Association for Computational Linguistics.

\bibitem[{Stab et~al.(2018)Stab, Miller, Schiller, Rai, and Gurevych}]{stab-etal-2018-cross}
Christian Stab, Tristan Miller, Benjamin Schiller, Pranav Rai, and Iryna Gurevych. 2018.
\newblock \href {https://doi.org/10.18653/v1/D18-1402} {Cross-topic argument mining from heterogeneous sources}.
\newblock In \emph{Proceedings of the 2018 Conference on Empirical Methods in Natural Language Processing}, pages 3664--3674, Brussels, Belgium. Association for Computational Linguistics.

\bibitem[{Stein et~al.(2021)Stein, Ajjour, El~Baff, Al-Khatib, and Cimiano}]{stein2021same}
Benno Stein, Yamen Ajjour, Roxanne El~Baff, Khalid Al-Khatib, and Philipp Cimiano. 2021.
\newblock \href {https://ceur-ws.org/Vol-2921/overview.pdf} {Same side stance classification}.

\bibitem[{Thomas et~al.(2006)Thomas, Pang, and Lee}]{thomas2006get}
Matt Thomas, Bo~Pang, and Lillian Lee. 2006.
\newblock Get out the vote: Determining support or opposition from congressional floor-debate transcripts.
\newblock In \emph{Proceedings of the 2006 Conference on Empirical Methods in Natural Language Processing}, EMNLP '06, page 327–335, USA. Association for Computational Linguistics.

\bibitem[{Tunstall et~al.(2022)Tunstall, Reimers, Jo, Bates, Korat, Wasserblat, and Pereg}]{tunstall2022efficient}
Lewis Tunstall, Nils Reimers, Unso Eun~Seo Jo, Luke Bates, Daniel Korat, Moshe Wasserblat, and Oren Pereg. 2022.
\newblock \emph{Efficient Few-Shot Learning Without Prompts}.

\bibitem[{van~der Meer et~al.(2022)van~der Meer, Reuver, Khurana, Krause, and Santamaria}]{van2022will}
Michiel van~der Meer, Myrthe Reuver, Urja Khurana, Lea Krause, and Selene~Baez Santamaria. 2022.
\newblock Will it blend? mixing training paradigms \& prompting for argument quality prediction.
\newblock In \emph{Proceedings of the 9th Workshop on Argument Mining}, pages 95--103.

\bibitem[{Van~Miltenburg et~al.(2021)Van~Miltenburg, van~der Lee, and Krahmer}]{van2021preregistering}
Emiel Van~Miltenburg, Chris van~der Lee, and Emiel Krahmer. 2021.
\newblock Preregistering nlp research.
\newblock In \emph{Proceedings of the 2021 Conference of the North American Chapter of the Association for Computational Linguistics: Human Language Technologies}, pages 613--623.

\bibitem[{Varadarajan et~al.(2022)Varadarajan, Soni, Wang, Luhmann, Schwartz, and Inoue}]{varadarajan-etal-2022-detecting}
Vasudha Varadarajan, Nikita Soni, Weixi Wang, Christian Luhmann, H.~Andrew Schwartz, and Naoya Inoue. 2022.
\newblock \href {https://doi.org/10.18653/v1/2022.nlpcss-1.16} {Detecting dissonant stance in social media: The role of topic exposure}.
\newblock In \emph{Proceedings of the Fifth Workshop on Natural Language Processing and Computational Social Science (NLP+CSS)}, pages 151--156, Abu Dhabi, UAE. Association for Computational Linguistics.

\bibitem[{Vu et~al.(2020)Vu, Wang, Munkhdalai, Sordoni, Trischler, Mattarella-Micke, Maji, and Iyyer}]{vu2020exploring}
Tu~Vu, Tong Wang, Tsendsuren Munkhdalai, Alessandro Sordoni, Adam Trischler, Andrew Mattarella-Micke, Subhransu Maji, and Mohit Iyyer. 2020.
\newblock Exploring and predicting transferability across nlp tasks.
\newblock In \emph{Proceedings of the 2020 Conference on Empirical Methods in Natural Language Processing (EMNLP)}, pages 7882--7926.

\bibitem[{Waldis et~al.(2024)Waldis, Hou, and Gurevych}]{waldis-etal-2024-dive}
Andreas Waldis, Yufang Hou, and Iryna Gurevych. 2024.
\newblock \href {https://aclanthology.org/2024.findings-eacl.146} {Dive into the chasm: Probing the gap between in- and cross-topic generalization}.
\newblock In \emph{Findings of the Association for Computational Linguistics: EACL 2024}, pages 2197--2214, St. Julian{'}s, Malta. Association for Computational Linguistics.

\bibitem[{Wei and Mao(2019)}]{wei2019modeling}
Penghui Wei and Wenji Mao. 2019.
\newblock Modeling transferable topics for cross-target stance detection.
\newblock In \emph{Proceedings of the 42nd International ACM SIGIR Conference on Research and Development in Information Retrieval}, pages 1173--1176.

\bibitem[{Williams et~al.(2018)Williams, Nangia, and Bowman}]{williams2018broad}
Adina Williams, Nikita Nangia, and Samuel~R Bowman. 2018.
\newblock A broad-coverage challenge corpus for sentence understanding through inference.
\newblock In \emph{Proceedings of NAACL-HLT}, pages 1112--1122.

\bibitem[{Wolf et~al.(2020)Wolf, Debut, Sanh, Chaumond, Delangue, Moi, Cistac, Rault, Louf, Funtowicz et~al.}]{wolf2020transformers}
Thomas Wolf, Lysandre Debut, Victor Sanh, Julien Chaumond, Clement Delangue, Anthony Moi, Pierric Cistac, Tim Rault, R{\'e}mi Louf, Morgan Funtowicz, et~al. 2020.
\newblock Transformers: State-of-the-art natural language processing.
\newblock In \emph{Proceedings of the 2020 conference on empirical methods in natural language processing: system demonstrations}, pages 38--45.

\bibitem[{Xu et~al.(2018)Xu, Paris, Nepal, and Sparks}]{xu2018cross}
Chang Xu, Cecile Paris, Surya Nepal, and Ross Sparks. 2018.
\newblock Cross-target stance classification with self-attention networks.
\newblock In \emph{Proceedings of the 56th Annual Meeting of the Association for Computational Linguistics (Volume 2: Short Papers)}, pages 778--783.

\bibitem[{Yang and Urbani(2021)}]{yang2021tribrid}
Song Yang and Jacopo Urbani. 2021.
\newblock Tribrid: Stance classification with neural inconsistency detection.
\newblock In \emph{Proceedings of the 2021 Conference on Empirical Methods in Natural Language Processing}, pages 6831--6843.

\end{thebibliography}

\section{Language Resource References}
\label{lr:ref}
\bibliographystylelanguageresource{lrec-coling2024-natbib}
\bibliographylanguageresource{languageresource}

\newpage

\appendix

\onecolumn
 \section{Results on Individual Datasets}\label{app:data_result}

Below are Table~\ref{tab:cross-topic-2} and \ref{tab:cross-topic-3+}, which describe results per individual dataset.

\begin{table*}[h!]
\centering
\small{
\begin{tabular}{llc|c|cccc|cc}
\hline
\textbf{Task ↓} &\multicolumn{2}{c}{\textbf{\small{Architecture ↓}}} & \textbf{Model ↓} & \multicolumn{4}{c|}{\textbf{datasets (2 labels)}}  &  \\ \hline
 &\multicolumn{2}{c}{} &  & \textbf{ibmcs} & \textbf{pers} & \textbf{argmin} & \textbf{scd} & \textbf{Mean (std)} \\ \hline

   Pro/Con   & \multicolumn{2}{c|}{cross-encoding} &  RoBERTa-large & .458 (.007) & .554 (.002) &.551 (.001) & .574 (.021) & .534 (.052) \\ 
 
  & & & +NLI & .498 (.07)  & .566 (.002) & .527 (.002) & .574 (.002) & .541 (.035)   \\ 
    & \multicolumn{2}{c|}{bi-encoding} & +STS & .520 (< .002) &.634 (< .001) & \textbf{.599 (.002)} & .461 (.003) & .554 (.078)\\ 
  & \multicolumn{2}{c|}{} & +NLI & .566 (.090) & .635 (.001) & .547 (.010)   & .578 (.003) & .582 (.038)\\ 

\hline
SSSC &\multicolumn{2}{c|}{cross-encoding} &   RoBERTa-large & .518 (.007) & .702 (.06) & .411 (.05) & .526 (.024) & .539 (.121) \\ 
   & \multicolumn{2}{c|}{} &  +NLI & \textbf{.734 (.004)} & \textbf{.766 (.001)} &  .564 (.003) & .428 (.002) &  \textbf{623 (.157)} & \\ 
    & \multicolumn{2}{c|}{bi-encoding} &   +STS & .474 (< .001) & .518 (< .001) & .447 (< .001) & .436 (< .001) & 469 (.0360) \\ 
& \multicolumn{2}{c|}{} & +NLI & .491 (< .001) & .528 (<.05) & .525 (< .001) & \textbf{.627} ( < .001) & .542 (.059)\\
\hline
\multicolumn{4}{c|}{dataset mean (std)} & .532 (.088) & \textbf{.613 (.088)} & .521 (.062) & .526 (.075) &   \\
\hline
\end{tabular}
}
\caption{Results (M F1 ) (macro) and SD over 5 random seeds) for stance datasets testing the cross-topic scenario with 2 labels in the Pro/Con task. All models were trained with the same hyperparameters (batch=2, LR =$16 \times 10^{-6}$ ,epochs=20), , and same N = 100 train split. Train/Test split as defined in Table~\ref{tab:datasets}}\label{tab:cross-topic-2}
\end{table*}

\begin{table*}[h!]
\centering
\small{
\begin{tabular}{llccc|c|ccccccccccccccc}
\hline
\textbf{Task ↓} && \multicolumn{2}{c}{\textbf{\small{Architecture ↓}}}  & &  \textbf{Model ↓} & \textbf{datasets (3 labels, cross-topic)} & & \\ \hline
 & \multicolumn{2}{c|}{} & \multicolumn{2}{c|}{} &  &\textbf{iac} & & \\ 
 \hline

\multicolumn{3}{c|}{Pro/Con}   & \multicolumn{2}{c|}{cross-encoding}  & RoBERTa-large & .345 (.008)  &  &\\ 
\multicolumn{3}{c|}{}   & \multicolumn{2}{c|}{} & +NLI & .345 (.001)   &\\ 
  \multicolumn{3}{c|}{}   & \multicolumn{2}{c|}{bi-encoding} & +STS & .280 (.039) & &\\ 
\multicolumn{3}{c|}{}   & \multicolumn{2}{c|}{} & +NLI &  .288 (.011) & & \\
\hline
 
\multicolumn{3}{c|}{SSSC} &\multicolumn{2}{c|}{cross-encoding} & RoBERTa-large & .389 (.020) \\ 
\multicolumn{3}{c|}{}   & \multicolumn{2}{c|}{} & +NLI & .455 (.013)  &\\ 

  \multicolumn{3}{c|}{}   & \multicolumn{2}{c|}{bi-encoding} & +STS & \textbf{.525 (.002)}  &\\ 
\multicolumn{3}{c|}{}   & \multicolumn{2}{c|}{} & +NLI & .516 (< .001) \\
\hline
\multicolumn{6}{c|}{dataset average} & .393 (.096) &  \\
\hline
\end{tabular}
}
\caption{Results (M and SD over 5 random seeds) for stance datasets testing the cross-topic scenario with more than 2 labels in the Pro/Con task definition. All models were trained with the same hyperparameters (batch=2, LR =$16 \times 10^{-6}$ ,epochs=20), and same N = 100 train split. Train/Test split as defined in Table~\ref{tab:datasets}.}\label{tab:cross-topic-3+}
\end{table*}

\begin{table*}[t]
\centering
\small{
\begin{tabular}{llcc|cc|cccccccccccccc}
\hline
\textbf{Task ↓} & {\small{\textbf{Architecture ↓}}}  & &  \textbf{Model ↓} & & \textbf{datasets (labels) (in-topic)}\\ \hline

&&  & &  \textbf{arc (4) } & \textbf{sem2016 (3)}  & \textbf{Mean (std)}\\ \hline

Pro/Con   & \multicolumn{2}{c|}{cross-encoding} & RoBERTa-large & .230 (< 007) & .443 (.002) & .337 (.151) \\ 
   & \multicolumn{2}{c|}{} & +NLI & .237 (.004) & .411 (.021) & .324 (.123)\\ 
  & \multicolumn{2}{c|}{bi-encoding} &  +STS & .223 (.001) & .492 (.004) &  .357 (.190) \\ 
 & \multicolumn{2}{c|}{} & +NLI &  .227 (< .001) & .389 (.02) & .308 (.115)  \\

\hline
SSSC &\multicolumn{2}{c|}{cross-encoding} &  RoBERTa-large & .477 (.017) &  .463 (.050) & .470 (.099)\\ 
& \multicolumn{2}{c|}{} & +NLI & \textbf{.547 (< .001)} & \textbf{.592 (.06)} & \textbf{.569 (.032)}\\ 

 & \multicolumn{2}{c|}{bi-encoding} &  +STS & .423 (.001) & .478 (< .001) & .450 (.039)\\ 
& \multicolumn{2}{c|}{} & +NLI & .431 (.007) & .460 (< .001) & .445 (.020)  \\
\hline 
\multicolumn{4}{c|}{dataset average} & .349 (.134) & \textbf{.466 (.061)}  \\
\hline
\end{tabular}
}

\caption{Results (in macro F1) for stance datasets testing the within-topic scenario with more than 2 labels in the Pro/Con task definition. All models were trained with the same hyperparameters (batch=2, LR =$16 \times 10^{-6}$ ,epochs=20), and train-test splits as defined in Table~\ref{tab:datasets}}\label{tab:in-topic-3+}
\end{table*}

\end{document}